\documentclass[letterpaper, 10 pt, conference]{ieeeconf}  

\IEEEoverridecommandlockouts                              
\usepackage[utf8]{inputenc}
\usepackage[T1]{fontenc}

\usepackage{graphicx}
\graphicspath{ {images/} }

\providecommand{\keywords}[1]{\textbf{\textit{Index terms---}} #1}
\begin{document}
\title{\LARGE \bf
Predicting Autism Spectrum Disorder Using Machine Learning Classifiers
}


\author{Koushik Chowdhury\\
\textit{MSc. Student} \\
\textit{Saarland University}\\
Saarbrücken, Germany \\
s8kochow@stud.uni-saarland.de

\and
Mir Ahmad Iraj\\
\textit{MSc. Student} \\
\textit{American International University-Bangladesh}\\
Dhaka, Bangladesh \\
13-23822-1@student.aiub.edu

}

\maketitle
\thispagestyle{empty}
\pagestyle{empty}

\begin{abstract}

Autism Spectrum Disorder (ASD) is on the rise and constantly growing. Earlier identify of ASD with the best outcome will allow someone to be safe and healthy by proper nursing. Humans can hardly estimate the present condition and stage of ASD by measuring primary symptoms. Therefore, it is being necessary to develop a method that will provide the best outcome and measurement of ASD. This paper aims to show several measurements that implemented in several classifiers. Among them, Support Vector Machine (SVM) provides the best result and under SVM, there are also some kernels to perform. Among them, the Gaussian Radial Kernel gives the best result. The proposed classifier achieves 95\% accuracy using the publicly available standard ASD dataset.

\end{abstract}

\section{INTRODUCTION}

Analyzing data for different purposes (like a prediction or measuring performances) is called the data mining process. It has several tasks such as association rule mining, classification, prediction and clustering. Researchers have found a suitable way of different purposes of data mining for different fields of research. Here in this paper, we have gone through to predict autism spectrum disorder (ASD) by applying the data mining technique specifically in Support-Vector Machine (SVM).

Autism spectrum disorder (ASD) is a neurological and developmental disorder that begins early in childhood and lasts throughout a person's life. It affects how a person acts and interacts with others. It is called a "spectrum" disorder because people with ASD can have a range of symptoms \cite{c1}. People with ASD might have problems talking with others, or they might not look in the eye when someone talks to them. They may also have restricted interests and repetitive behaviors. Research says that both genes and environment play important roles and there is currently no standard treatment for ASD but there are many ways to increase a person's ability to grow normally. We really need to check a person’s behavior and symptoms whether he is autistic or not. Therefore, we need a strong dataset to perform a technique that will identify a person’s autism spectrum disorder. However, very limited autism datasets associated with clinical or screening are available and most of them are genetic in nature.

A suitable data set of autistic spectrum disorder (ASD) has been selected that is combined with three different data sets such as ASD Screening Data for Child, ASD Screening Data for Adolescent and ASD Screening Data for Adult \cite{c1}. We have merged three of these data sets into one dataset for our mission. Our desired dataset consists of ten individual characteristics and ten behavioral features in binary data. We consider with their data type respectively such as String, Number, Boolean and the ten behavioral questions is in Binary data type. Using Python programming language and python libraries such as sklearn, pandas, keras, numpy and matplotlib libraries, we have got our predicting results both in graphical and numeric from SVM.

\section{Background Study}

\subsection{Literature Review} 
Analysis of the increasing trend of using modern machine learning as well as data mining technologies to explore data efficiently is on the rise. We are at the age where we need more than perfection.

Some researchers used machine learning techniques to measure data in the most accurate form. Implementing datasets in several classifiers and several algorithms, the outcome was in terms of calculating the accuracy rate. Such as, research was in 2016 for breast cancer risk prediction and diagnosis by using machine learning algorithms \cite{c2}. There they had performed a comparison between different machine learning algorithms. Such as Support Vector Machine (SVM), Decision Tree (C4.5), Naive Bayes (NB) and k Nearest Neighbors (k-NN). These algorithms had been implemented on the Wisconsin Breast Cancer datasets \cite{c2}. Their experimental results showed that SVM gave the highest accuracy with the lowest error rate and all experiments were executed within a simulation environment and conducted in WEKA data mining tool \cite{c2}.

There are many more similar researches took place on several topics and several datasets. Here, we would like to mention another one that was on Magnetic Resonance Imaging (MRI) stroke classification by Support Vector Machine (SVM) \cite{c3}. They performed a method to classify the MRI images of the brain related stroke. The MRI images for stroke classification used Gabor filters and Histograms to extract features from the images and the features were classified using Support Vector Machine (SVM) with various kernels. Finally, the experimental results were shown that the presented method achieves satisfactory classification accuracy and the classification accuracy was the best for the linear kernel.

From this kind of research mentioned above here, we gather knowledge and related information that we have used for our work. We aim to improve the result of the prediction of the ASD Screening test of an individual by doing some analysis of Data Mining techniques.

\subsection{Data Understanding}
\subsubsection{Attributes} 
Our work uses publicly available standard data sets \cite{c4}. Score such as A1\_Score, A2\_Score are the result of a questionnaire survey by the Autism Research Center at the University of Cambridge, UK \cite{c5}. The Dataset consists of nine individual characteristics and ten behavioral features. They are following.
 \begin{center}
        \begin{tabular}{|c| c |}
       
        \hline
         \textbf{Attribute} &\textbf{Type}\\\hline
         
         gender  &String  \\\hline
         ethnicity &String  \\\hline
         jaundice &Boolean (yes or no) \\\hline
         PDD &Boolean (yes or no)\\\hline
         relation &String\\\hline
         country\_of\_res &String\\\hline
         did\_the\_qn\_before &Boolean (yes or no)\\\hline
         age\_desc &Integer\\\hline
         A1\_Score &Binary (0, 1)\\\hline
         A2\_Score &Binary (0, 1)\\\hline
         A3\_Score &Binary (0, 1)\\\hline
         A4\_Score &Binary (0, 1)\\\hline
         A5\_Score &Binary (0, 1)\\\hline
         A6\_Score &Binary (0, 1)\\\hline
         A7\_Score &Binary (0, 1)\\\hline
         A8\_Score &Binary (0, 1)\\\hline
         A9\_Score &Binary (0, 1)\\\hline
         A10\_Score &Binary (0, 1)\\\hline
         Class/ASD &Boolean (yes=1 or no=0)\\\hline

        \end{tabular}
\end{center}
\subsubsection{Missing Value}
The dataset contains so many missing values. Dealing with these missing values was the biggest challenge as we have 19 variables. There are a few ways to handle the missing values. For example, replace the missing values with averaged values or delete instances that have missing values. Since we have 19 variables, we decide to remove the instances with missing values.
\section{Approaches}
\subsection{Data Classification}
There are two steps in data classification. The first step is known as a learning step. A given set of classes based on the analysis of a set of data instances is explained. Each instance belongs to a predefined class. In the last step, the data set is tested using various machine learning techniques that are used to calculate the classification accuracy, AUC value, precision, recall, etc. A model is then designed that predicts the future outcome based on the historical or recorded data. There are various machine learning techniques for classification.
We have used the following machine learning classifier in our work, which is seen in the section 'Classifiers'.
\subsection{Classifiers}
We applied the following machine learning classifier to the data set. First, we divided our data set into two categories: training (70\%) and testing (30\%). After splitting the data set, we used this machine learning classifier to determine the results of the assessment matrices.
\begin{itemize}
    \item Naïve Bayes.
    \item K-Nearest Neighbor(kNN).
    \item Logistic Regression.
    \item Gradient Boosting.
    \item Support Vector Machine.
    \item Decision Tree.
    \item MLP Classifier.
\end{itemize}

\subsection{Evaluation Metrics} 
We select the following evaluation metrics to evaluate our results. These metrics, especially accuracy and AUC value, help us to find the best classifier for the data set.
\begin{itemize}
    \item Accuracy.
    \item AUC Value.
    \item Precision.
    \item Recall.
    \item F1 Score.
\end{itemize}

\section{Results and Finding}
\subsection{Comparison between different classifiers} 

In order to prove which classifier is performing better than others, we must need a comparison between them.
\subsubsection{Results} To compare between classifiers we need proper measurement result of their accuracy rate, AUC value, Precision, Recall and F1-Score.  So they are..
\\
\begin{table}[hbt!]
        \begin{tabular}{| c| c |c| c| c| c| c|c|}
       
        \hline
         & NB  &  kNN    &  LR & GB & SVM &DT &MLP\\\hline
         
         Acc &0.76 &0.92 & 0.915 &0.921 &0.95 & 0.87 & 1\\\hline
         AUC &0.78 &0.81 &0.90& 0.88 &0.87 &0.86 &0.67\\\hline
         Pre (no) &0.86 &0.87 &0.93 &0.89 &0.89 &0.90 &1\\\hline
         Pre (yes) &0.69 &0.73 &0.89 &0.92 &0.91 &0.84 &0.47\\\hline
         Rec (no) &0.81 &0.84 &0.94 &0.96 &0.95 &0.92 &0.37\\\hline
         Rec (yes) &0.76 &0.78 &0.87 &0.80 &0.79 &0.82 &1\\\hline
         F1 (no) &0.83 &0.86 &0.93 &0.93 &0.92 &0.91 &0.54\\\hline
         F1 (yes) &0.72 &0.75 &0.88 &0.85 &0.84 &0.83 &0.64\\\hline

        \end{tabular}
        \caption{Classifiers Results}
\end{table}

From the table I, Naïve Bayes is providing the lowest accuracy that is 0.76 and MLP Classifier is showing the highest accuracy that is 1.00. But MLP Classifier is providing the lowest AUC value which is only 0.686. Moreover, comparing with MLP Classifier’s AUC value, Precision, Recall, F1-score are not even close to other classifiers. MLP Classifier gets the data set over-fit. This can be a cause of MLP Classifier’s not working well with small datasets. So if we don’t count MLP Classifier in that way, Support Vector Machine (SVM) would be the best performer for this kind of operation as SVM has the second-best accuracy rate that is 0.950 which is 95\%. Though Logistic Regression is providing the highest AUC value that is 0.903, 
The Support Vector Machine shows a decent overall result for all evaluation metrics. So comparing these sides, we found support Vector Machine is very much appropriate for this kind of operation.

\subsubsection{ROC Curve}
The ROC curve is a graphical diagram that illustrates the probability curve of binary classifiers. 
 \begin{figure}[hbt!]
 \centering
     \includegraphics[scale=0.7]{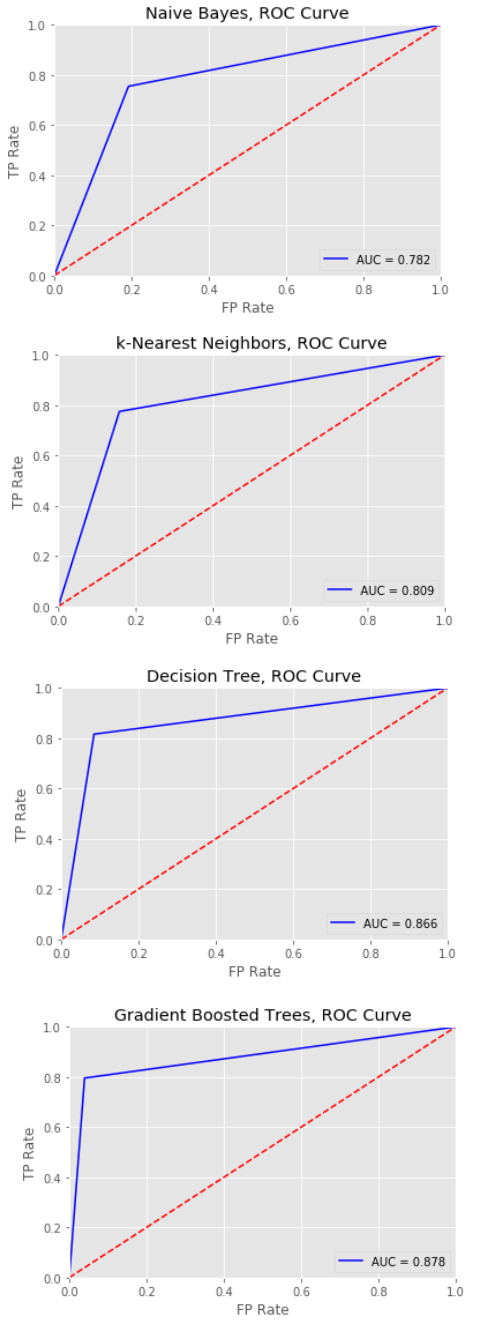}
      \caption{ROC Curve of Classifiers}
      \label{figurelabel}
   \end{figure}
Here, from figure 1, we can see that there are four ROC curves of four classifiers and they are plotted according to their AUC value. A perfect test result has an AUC value of 1.0, whereas random chance gives an AUC value of 0.5 or may lower. Here, our desired classifiers have AUC value of 0.782 Naïve Bayes, 0.809 k-Nearest Neighbor, 0.866 Decision Tree, 0.878 Gradient Boosted Trees. Gradient Boosting gives the best AUC value if we only compare these 4 classifiers. This is the advantage of ROC curve measurement that we can compare two or more sensitive tests visually and simultaneously in one figure. 

Now we want to compare between Logistic Regression, SVM and MLP Classifier. According to figure 2, Logistic Regression gives the best result here in the ROC curve and MLP Classifier gives the worst result here and SVM gives the best appreciable result in accuracy. 
 \begin{figure}[hbt!]
 \centering
     \includegraphics[scale=0.7]{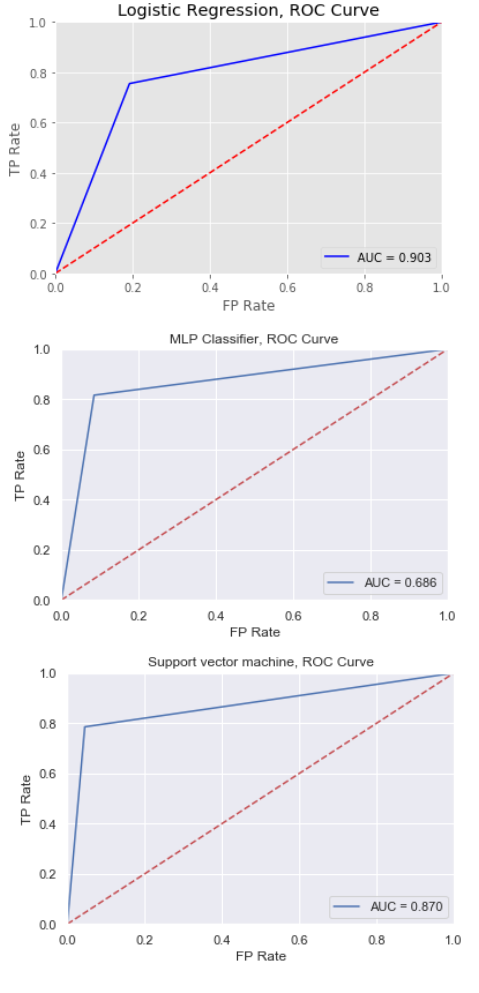}
      \caption{ROC Curve of Classifiers}
      \label{figurelabel}
   \end{figure}

From the above figure 2 we can see there are three ROC curves plotted according to their AUC values. To compare between them we need to check out their AUC values. Logistic Regression provides AUC of 0.903, MLP Classifier 0.686 and SVM provides AUC of 0.870. Here in ROC curve Logistic Regression gives the best result and MLP Classifier gives the worst. SVM is slightly less than Logistic Regression here in ROC curves but it gives the best appreciable result in accuracy. 

\subsection{Findings}

MLP Classifier provides 100\% accuracy but its AUC value, Precision, Recall and F1-Score are very low. So it is not fit for our dataset. Logistic Regression and SVM provide almost same values and they are very much fit for this kind of dataset. SVM works slightly better than Logistic Regression if we evaluate based on their accuracy metrics. So it is better to select SVM for this kind of datasets.

\subsection{Prediction using Support Vector Machine}
Support Vector Machine algorithm uses several sets of mathematical functions that are defined as SVM kernels. The function of these SVM kernels is to take data as input and transform it into the required form. There are different kinds of SVM algorithms that use different types of kernel functions and these functions can be different types \cite{c6}. Most used SVM kernels are Linear SVM kernel, Polynomial SVM Kernel, Gaussian Radial Basis Kernel and Sigmoid SVM kernel. Our data does not suit the linear kernel since one dependent response and one or more independent variables are linear concerns. Out data have 2 dependent responses, they are Yes and No. But we continue our operation by implementing in rest of the most used three SVM kernels.

\subsubsection{SVM Kernel Results}
We need to find which SVM kernel is performing the best with our dataset. So we have to compare their measurement result of accuracy rate, AUC value, Precision, Recall and F1-Score.  
\begin{table}[hbt!]
\centering
        \begin{tabular}{| c| c |c| c|}
       
        \hline
         & Polynomial  &  Gaussian    &  Sigmoid\\\hline
         
         Acc &0.9479 &0.9503 &0.4570 \\\hline
         AUC &0.8318 &0.8703 &0.4015 \\\hline
         Pre (no) &0.87 &0.89 &0.58\\\hline
         Pre (yes) &0.81 &0.91 &0.22 \\\hline
         Rec (no) &0.90 &0.95 &0.60 \\\hline
         Rec (yes) &0.77 &0.79 &0.60 \\\hline
         F1 (no) &0.89 &0.92 &0.59 \\\hline
         F1 (yes) &0.79 &0.84 &0.59 \\\hline

        \end{tabular}
        \caption{SVM Kernel Results}
\end{table}
Here from the above comparison table we can see that Sigmoid SVM kernel is providing very poor result and it is always less than half (0.5). On the other hand Polynomial Kernel and Gaussian Radial Basis Kernel are performing well. Among them Gaussian Radial Basis Kernel is proving better result than Polynomial Kernel.

\subsubsection{ROC Curves of SVM Kernels}
Here the ROC Curves of SVM Kernels are plotted according to their AUC value. A perfect test result has an AUC value of 1.0, whereas random chance gives an AUC value of 0.5 or may more lower. 

Here, from figue 3, we can see AUC value of SVM Polynomial kernel is 0.832, SVM Gaussian Radial Basis kernel is 0.870 and SVM Sigmoid Kernel is 0.401. So here SVM Sigmoid kernel performance is very poor but Polynomial kernel and Gaussian Radial Basis kernel is performed well almost similar.

\begin{figure}[hbt!]
 \centering
     \includegraphics[scale=0.7]{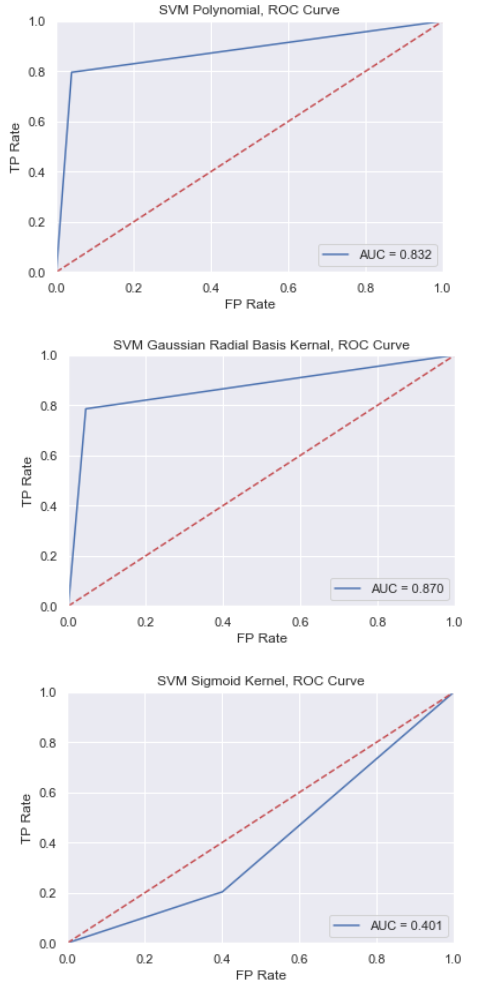}
      \caption{ROC Curve of SVM Kernels}
      \label{figurelabel}
   \end{figure}

\subsubsection{SVM Finding}
SVM Sigmoid kernel is giving very poor result for this dataset, whereas Gaussian Radial Basis kernel and Polynomial kernel performed almost similar and that is good for our dataset. But SVM Gaussian Radial Basis kernel slightly performed better than SVM Polynomial Kernel. So we can say that SVM Gaussian Radial Basis kernel is the best performed algorithm as well as classifier for this kind of datasets. The following figure 4 shows a sample comparison of the actual results and the results of SVM kernels.

\begin{figure}[hbt!]
\centering
     \includegraphics[scale=0.7]{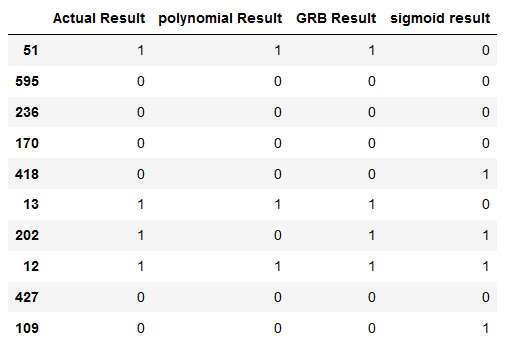}
      \caption{Actual Results vs Predicted Results (sample comparison)}
      \label{figurelabel}
   \end{figure}

\section{Future Work}
We all know that, with the arrival of the massive cloud age, big data mining is the latest research trend. In the future, to focus on the study of achieving efficient and accurate classification under the environment of data mining or machine learning, we need to work with a huge number of data. Also, our suggestion is to do Deep Learning instead of using traditional classifiers.
\section{Conclusion}
This work investigates the efficiency of various kinds of machine learning classifiers for a specific kind of dataset. We choose medical data to predict autism spectrum disorder. To perform the work smoothly, we mapped the features and build the dataset free of missing values. We found that the Support Vector Machine ( SVM) is very suitable for this dataset based on the experiments conducted in this study, and it performed very well. So we choose SVM to analyze the dataset more deeply. To get a more accurate result, we implemented the most used kernels of SVM. Among them, we found SVM Gaussian Radial Basis kernel is the best performed algorithm as well as a classifier for this kind of dataset. To deal with such a kind of medical dataset and to find the most powerful classifier that gives the most reliable result was a great challenge.


\end{document}